\documentclass[lettersize]{IEEEtran}
\usepackage{times}
\usepackage{amsmath,amsfonts,amssymb}
\usepackage[noadjust]{cite}
\usepackage{notoccite}
\usepackage{amsfonts} 
\usepackage[ruled,vlined,linesnumbered]{algorithm2e}
\usepackage{multicol}
\usepackage[dvipsnames]{xcolor}
\usepackage[bookmarks=true,hidelinks]{hyperref}
\usepackage{svg}
\usepackage{subcaption}
\usepackage{multirow}
\usepackage[font=small,skip=2pt]{caption}
\usepackage{float}
\usepackage{diagbox}
\usepackage{bm}
\usepackage{changepage}
\usepackage{stfloats}
\usepackage[english]{babel}

\let\oldnl\nl
\newcommand{\nonl}{\renewcommand{\nl}{\let\nl\oldnl}}

\newcommand{\edit}[1]{\textcolor{black}{#1}}
\begin{document}
\title{Learning Prehensile Dexterity by Imitating and Emulating State-only Observations}
\author{
Yunhai Han$^{1}$, Zhenyang Chen$^{1}$,  Kyle A Williams$^{1}$, Harish Ravichandar$^{1}$
\thanks{$^{1}$
Georgia Institute of Technology, Atlanta, GA 30332, USA (email: \{yhan389, zchen927, kwilliams319,   harish.ravichandar\}@gdatech.edu).
}
}
\maketitle
\begin{abstract}
When human acquire physical skills (e.g., tool use) from experts, we tend to first learn from merely observing the expert. But this is often insufficient. We then engage in practice, where we try to emulate the expert and ensure that our actions produce similar effects on our environment. 
Inspired by this observation, we introduce \textit{\textbf{C}ombining \textbf{IM}itation and \textbf{E}mulation for Motion \textbf{R}efinement} (CIMER) -- a two-stage framework to learn dexterous prehensile manipulation skills from \textit{state-only} observations. 
CIMER's first stage involves \textit{imitation}: simultaneously encode the complex interdependent motions of the robot hand and the object in a structured dynamical system. This results in a reactive \textit{motion generation} policy that provides a reasonable \textit{motion prior}, but lacks the ability to reason about contact effects due to the lack of action labels. 
The second stage involves \textit{emulation}: learn a \textit{motion refinement} policy via reinforcement that adjusts the robot hand's motion prior such that the learned \textit{object} motion is reenacted.
CIMER is both \textit{task-agnostic} (no task-specific reward design or shaping) and \textit{intervention-free} (no additional teleoperated or labeled demonstrations).
Detailed experiments with prehensile dexterity reveal that 
i) imitation alone is insufficient, but adding emulation drastically improves performance, 
ii) CIMER outperforms existing methods in terms of sample efficiency and the ability to generate realistic and stable motions,
iii) CIMER can either zero-shot generalize or learn to adapt to novel objects from the YCB dataset, even outperforming expert policies trained with action labels in most cases. Source code and videos are available at \href{https://sites.google.com/view/cimer-2024/}{https://sites.google.com/view/cimer-2024/}.

\end{abstract}

\section{Introduction}
\edit{Learning dexterous manipulation skills involving multi-finger hands (e.g., relocation and tool use)  presents numerous challenges, such as high-dimensional state and action spaces, complex nonlinear dynamics, and contact effects.} Unfortunately,  both imitation learning (IL) and reinforcement learning (RL) tend to struggle with such complex tasks when employed alone. Specifically, IL suffers from distribution shift~\cite{ross2011reduction, ravichandar2020recent} and RL from poor sample efficiency~\cite{li2023deep}. To address these challenges, prior works have shown promise in leveraging demonstrations to expedite reinforcement learning~\cite{Rajeswaran-RSS-18, qin2022dexmv, haldar2023teach, xu2023dexterous, kannan2023deft, mandikal2021learning}.

However, collecting demonstrations for dexterous manipulation in the form of state-action pairs can be challenging and cumbersome due to complex system design and setup \cite{kumar2015mujoco,handa2020dexpilot}.
To alleviate this burden, recent research has focused on learning dexterous manipulation skills from state-only observations~\cite{chen2022dextransfer, radosavovic2021state, shaw2023videodex}.
However, learning without action labels can result in kinematic motions that are oblivious to force and contact. This lack of knowledge is particularly limiting in dexterous \textit{prehensile} manipulation tasks (e.g., grasping and tool use) as they exhibit heightened sensitivity to applied force~\cite{chen2022dextransfer, bicchi1995closure, zheng2005coping, zhu2003synthesis}. 
To compensate, existing methods often require additional teleoperated demonstrations~\cite{shaw2023videodex, arunachalam2023dexterous}, human-in-the-loop corrections~\cite{wang2024dexcap}, task-specific rewards~\cite{qin2022dexmv, radosavovic2021state, qin2022from, she2022learning}, or user-defined sub-goal images~\cite{xu2023dexterous, hu2023reboot}. 

\begin{figure}[t]
\centering
\includegraphics[width=\linewidth]{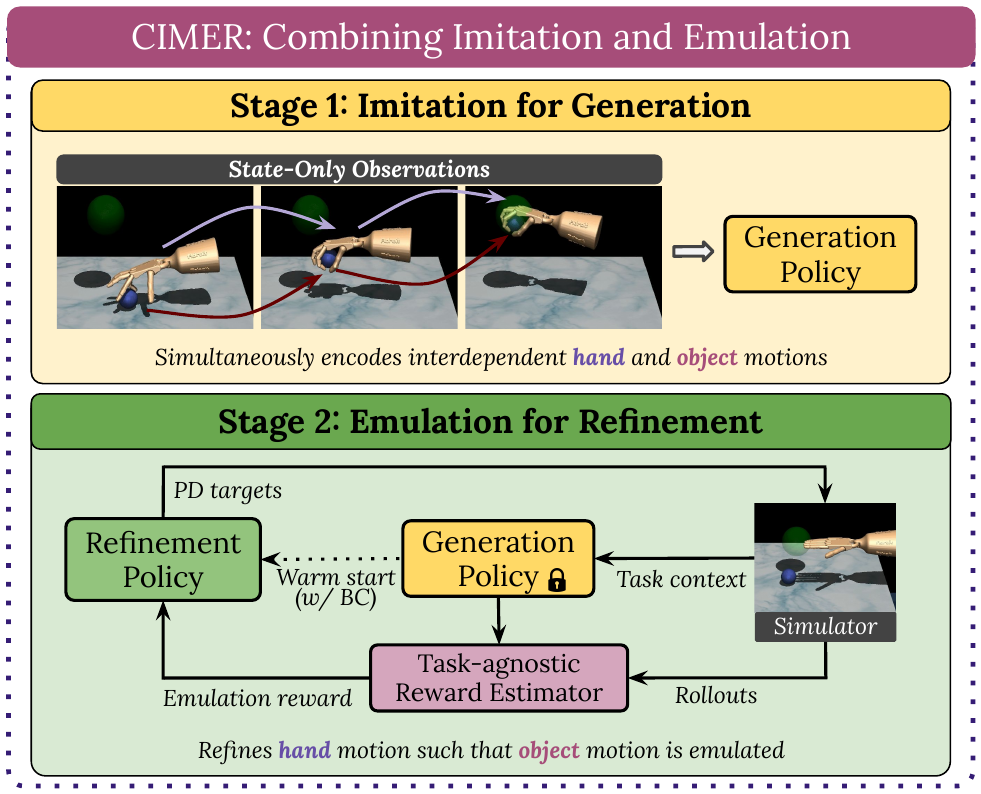}
\caption{CIMER is a \textit{task-agnostic} and \textit{intervention-free} framework to learn dexterous manipulation skills from state-only observations by learning to first generate interdependent desired motions of the robot hand and the object (\textit{Imitation}), and then refine the generated robot motion in order to reenact the learned object motion (\textit{Emulation}).
}
\label{fig:block_diagram}
\end{figure}

In this work, we contribute a novel two-stage framework to 
learn dexterous prehensile manipulation skills from
state-only observations. We call our framework CIMER, short for \textit{\textbf{C}ombining \textbf{IM}itation and \textbf{E}mulation for Motion \textbf{R}efinement} (see Fig.~\ref{fig:block_diagram}). 
The first stage is \textit{Imitation}: CIMER encodes the interdependent desired motions of both the robot hand and the object from the observations into a structured time-invariant dynamical system that acts as a reactive policy and a reasonable \textit{motion prior} (see Sec.~\ref{subsec:imitation}).
\edit{The second stage is \textit{Emulation}: CIMER learns to refine the \textit{robot's} motion prior based on context by optimizing a \textit{task-agnostic} reward that incentivizes the reenactment of the learned \textit{object} motion~\cite{schaal1999imitation}.}


\edit{Three key insights motivate our design of CIMER. 
First, dexterous prehensile manipulation involves complex and inherently \textit{interdependent} motions of the robot hand and the object. As such, CIMER's imitation stage captures both their motions simultaneously using a single but structured dynamical system.
Second, state-only observations of the expert unambiguously demonstrate how the \textit{object} is supposed to move, and contain valuable (albeit crude) information about the robot hand's motion. 
As such, CIMER learns to only refines the hand motion and relies on the learned object motion for reward signals.
Third, separating motion generation and refinement amounts to first learning what the robot must do from \textit{state-only} observations, and then learning how to do it via \textit{self-guided} experimentation. 
Unlike prior methods, CIMER is both \textit{intervention-free} and \textit{task-agnostic} -- it neither requires in-context additional demonstrations nor task-specific reward design. See Sec.~\ref{sec:related_work} for a detailed discussion of related work.}
To both experimentally investigate the central issues and to evaluate CIMER, we conducted a series of thorough experiments on three dexterous prehensile manipulation tasks: \textit{Tool Use}, \textit{Object Relocation}, and \textit{Door Opening}. 

First, we demonstrate that pure imitation (without emulation) is rarely successful, and necessitates meticulous expert tuning of the low-level controller when it does.
But combining imitation and emulation (i.e., CIMER) significantly improves the task success rate, even while using the \textit{same} controller across tasks. We also uncover qualitative insights into how CIMER refines motions and why they succeed. 

Second, we demonstrate that CIMER tends to achieve better sample efficiency compared to existing methods that can learn from state-only observations.
Moreover, we highlight CIMER's ability to effectively leverage motion priors (encoded in a dynamical system) to produce realistic and stable motions, in contrast to baselines that tend to exploit the simulator and generate aggressive behaviors. We also show that policies trained using CIMER exhibit better robustness against changes to physical parameters such as mass and damping.

Third, we show that CIMER significantly outperforms the baselines in terms of zero-shot generalization to 17 unseen objects from the YCB dataset~\cite{calli2015benchmarking}. Notably, CIMER surpasses even the expert policy trained with action labels in many instances.
When CIMER can't readily generalize to a novel object with significant differences, it can effectively leverage previously learned skills to expedite adaptation. 

In summary, we contribute CIMER -- a two-stage framework that effectively combines imitation and emulation in order to learn dexterous prehensile manipulation skills from state-only observations. \edit{CIMER first learns a reasonable motion prior by imitating observations without relying on action labels or user interventions, and then employs a task-agnostic and self-guided strategy to learn how to refine the \textit{hand's} motion so that it emulates the learned \textit{object} motion.} 


\section{Related Work}\label{sec:related_work}
In this section, we discuss our contributions within the context of different related bodies of work. 

\edit{
\noindent\textbf{Learning dexterity from observations:}
Given the abundance of video data on the internet and the advances in motion retargetting (e.g.,~\cite{qin2022dexmv}), recent research has focused on learning dexterous manipulation skills from videos or state-only observations in general.}
A key challenge in learning from videos or state-only observations is the lack of action labels that speaks to the contact effects between the robot, the object, and the environment. Existing works circumvent this challenge in one of two ways. Some methods provide robots an opportunity to learn from interactions, either in simulation~\cite{qin2022dexmv, mandikal2021learning, dasari2023learning, agarwal2023dexterous} or in the real world~\cite{xu2023dexterous, kannan2023deft, hu2023reboot}. Other methods leverage additional in-domain teleoperated demonstrations from an expert~\cite{haldar2023teach, shaw2023videodex, arunachalam2023dexterous, qin2022from, arunachalam2023holo}. 
\edit{
Our work falls under the first category since it does not rely on any additional demonstrations. 
Unlike CIMER, many of the prior methods in the first category are limited to replicating a specific hand motion from a video clip, such as object grasping \cite{kannan2023deft, mandikal2021learning, dasari2023learning, agarwal2023dexterous}, bagel flipping \cite{haldar2023teach} or bottle opening \cite{arunachalam2023holo}. While the remaining methods are capable of learn a variety of skills and can handle changes to the objects' initial and target configurations, they require at least one of the following: task-specific reward engineering~\cite{qin2022dexmv, qin2022from}, user-provided sub-goal images as reward signals~\cite{xu2023dexterous, hu2023reboot}, and bespoke refinements tailored to dexterous grasping~\cite{chen2022dextransfer, pmlr-v205-chen23b}.
In contrast, CIMER can learn a variety of skills requiring neither task-specific reward shaping nor expert interventions.} 


\noindent
\edit{
\textbf{Learning dexterous teleoperation:} 
Another class of existing methods
enable a human operator to seamlessly teleoperate a dexterous robotic hand by leveraging learning to close the embodiment gap~\cite{handa2020dexpilot, qin2023anyteleop, li2019vision, sivakumar2022robotic}.}
However, it is important to note that these methods do \textit{not} learn autonomous policies.
Instead, they rely on real-time corrections and guidance from the human operator who tends to compensate for the robot's inability to reason about contact effects~\cite{Rui2022Feedback}.


\noindent\textbf{Learning low-dim skills from videos}: 
There is a large body of work that focuses on learning polices for low-dimensional manipulators (serial-link robots with parallel jaw grippers) from videos. One group of methods utilize a hierarchical framework to learn task-specific waypoint trajectories from human videos that can then be tracked using standard controllers~\cite{liu2018imitation, sharma2019third, xiong2021learning, bharadhwaj2023zero}.
Other methods learn a high-level planner that operates in a latent space and then use it to condition a low-level planner that can produce diverse robot behaviors~\cite{lynch2020learning, wang2023mimicplay, xu2023xskill}. However, unlike the low-dimensional skills learned by these methods, dexterous manipulation with multi-fingered hands inherently involves more complex finger-object interactions and contact effects. Further, similar to their counterparts that learn dexterous skills, many of these methods also rely on additional teleoperated demonstrations.
\noindent\textbf{Learning locomotion from observations}: 
Our approach to emulation is inspired by research in learning locomotion skills by imitating animals~\cite{peng2020learning} or animated characters~\cite{peng2018deepmimic, zhang2023vid2player3d}. These methods first extract the reference motion from video clips, and refine the learned motions to account for foot-ground contact effects and to ensure that the resulting gaits closely track the demonstrated ones.
However, merely imitating the observed robot motion is unlikely to succeed in dexterous manipulation since it is object-centric (see Sec.~\ref{subsec:why_emulation} for empirical validation). 
In contrast, we emphasize the reenactment of the learned \textit{object} motion.

\section{Approach}
We begin by formulating the problem of learning dexterous manipulation skills from state-only observations.

\subsection{Problem Formulation}
\label{subsec:formulation}
Let $\mathcal{D} = [\{{\mathrm{h}}_t^{(1)}, {\mathrm{o}}_t^{(1)}\}^{t=T^{(1)}}_{t=1}, \cdots, \{{\mathrm{h}}_t^{(N)},{\mathrm{o}}_t^{(N)}\}^{t=T^{(N)}}_{t=1}]$ denotes a dataset of $N$ state-only observations of a prehensile manipulation skill, where ${\mathrm{h}}_t^{(n)} \in \mathcal{H} \subseteq \mathbb{R}^n$ and ${\mathrm{o}}_t^{(n)} \in \mathcal{O} \subseteq \mathbb{R}^m$ respectively denote the robot hand state and object state (e.g., hand joint positions and object 6D poses) at time $t$ of $n$-th observation. We focus on prehensile manipulation skills since they are particularly challenging to learn without action labels. 
(see Sec. \ref{subsec:why_emulation}).
Our goal is to learn a policy from the dataset $\mathcal{D}$ that will enable the robot to autonomously perform the associated dexterous skill. In addition to the dataset, we assume access to a simulator. But, to ensure a \textit{task-agnostic} solution, we do not allow task-specific reward design or shaping. To remain \textit{intervention-free}, we do not assume access to additional labeled demonstrations. As such, our problem requires an approach that learns a robust dexterous manipulation skill purely from state-only observations and self-guided interactions with a simulator.

\subsection{Solution Overview}
\label{subsec:overview}
Our framework to address the above problem, Combining IMitation and Emulation for Motion Refinement (CIMER), operates in two stages: i) \textit{Imitation}: learn a \textbf{Motion Generation Policy} $\Phi$ in the form of a structured dynamical system that encodes the desired reference motions for both the robot and the object from state-only observations, and ii) \edit{\textit{Emulation}: learn a \textbf{Motion Refinement policy} $\Psi$ that compensates for the lack of action information by refining the desired \textit{robot} motions such that the learned \textit{object} motion is achieved.}
The overall pseudo-code is given in Appendix~ \ref{appendix:Pseudo_code}.

\subsection{Motion Generation via Imitation}
\label{subsec:imitation}
We view the complex interdependent desired motions of the robot and the object as solutions to a learnable underlying nonlinear behavioral dynamical system~\cite{ravichandar2020recent,pmlr-v205-xie23a, bahl2020neural}. \edit{Once learned, these dynamical systems can be integrated to predict the next desired robot state from initial conditions or the current state~\cite{ravichandar2020recent}.} 
In recent work, we developed a framework named KODex~\cite{han2023utility} that learns such underlying dynamics in a highly computationally-efficient manner using Koopman operator theory. \edit{Data-driven Koopman-based approaches help effectively encode highly nonlinear dynamical systems in a linear system using lifted global linearization~\cite{williams2015data,mauroy2020koopman}.}
Building on this, CIMER employs a Koopman-based method theory to learn the Motion Generation Policy $\Phi$ from the dataset $\mathcal{D}$. Note that KODex requires access to action labels, but CIMER does not.
\edit{Formally, we formulate the $\Phi$ as a \textit{linear} dynamical system in the \textit{lifted} space: $\phi({\mathrm{h}}_{t+1}, {\mathrm{o}}_{t+1}) = \mathbf{K}\ \phi({\mathrm{h}}_{t}, {\mathrm{o}}_{t})$, where $\phi: \mathbb{R}^{n+m} \rightarrow \mathbb{R}^{p}$ (with $p>(n+m)$) is a user-defined function that ``lifts" the robot hand and object states to a higher-dimensional space in which the underlying dynamics appear linear, and $\mathbf{K}$ is the Koopman matrix~\cite{williams2015data,mauroy2020koopman}}. We use a \textit{task-agnostic} second-order polynomial lifting function for all tasks in all our experiments. \edit{A key benefit of using a Koopman-based approach is that we can \textit{analytically} compute $\mathbf{K}$ (and thereby our motion generation policy $\Phi$) from $\mathcal{D}$ using least squares~\cite{mauroy2020koopman}}. To  extract the original states without a decoder, we augment the original states to the lifted states before learning the Koopman operator. 
See Appendix~\ref{appen:detail_kodex} for details. 
\edit{We then use the learned dynamical system as an autonomous policy to predict desired hand and object motions for any initial condition.}

\subsection{Motion Refinement via Emulation}
\label{subsec:emulation}
\edit{
Merely tracking the hand trajectory generated by the motion generation policy $\Phi$ does not guarantee task success in dexterous prehensile manipulation tasks due to the small margin for error and the lack of reasoning about contact effects
(see Sec.~\ref{subsec:why_emulation} for empirical validation). As such, CIMER refines the robot's \textit{hand} motion so as to ensure emulation of the learned \textit{object} motion. 
Specifically, CIMER learns $\Psi$ to refine finger and palm trajectories (as opposed to all the DoFs) generated by $\Phi$ such that the robot reenacts the object motion generated by $\Phi$.}
Interacting with the object helps compensate for the lack of action labels and account for contact effects. Once learned, CIMER employs a \textit{task-agnostic} PD controller to track the refined hand motions.

\edit{
We formulate emulation as an RL problem and maximize rewards computed from policies learned via imitation. 
Our task-agnostic reward function captures the essence of emulation: reenacting the desired object motion rather than only the expert's motion~\cite{schaal1999imitation}.}
To reduce training variance, we use Generalized Advantage Estimator (GAE) \cite{schulman2015high} and PPO \cite{schulman2017proximal}. 
\begin{figure}
     \centering
    \includegraphics[width=1\linewidth]{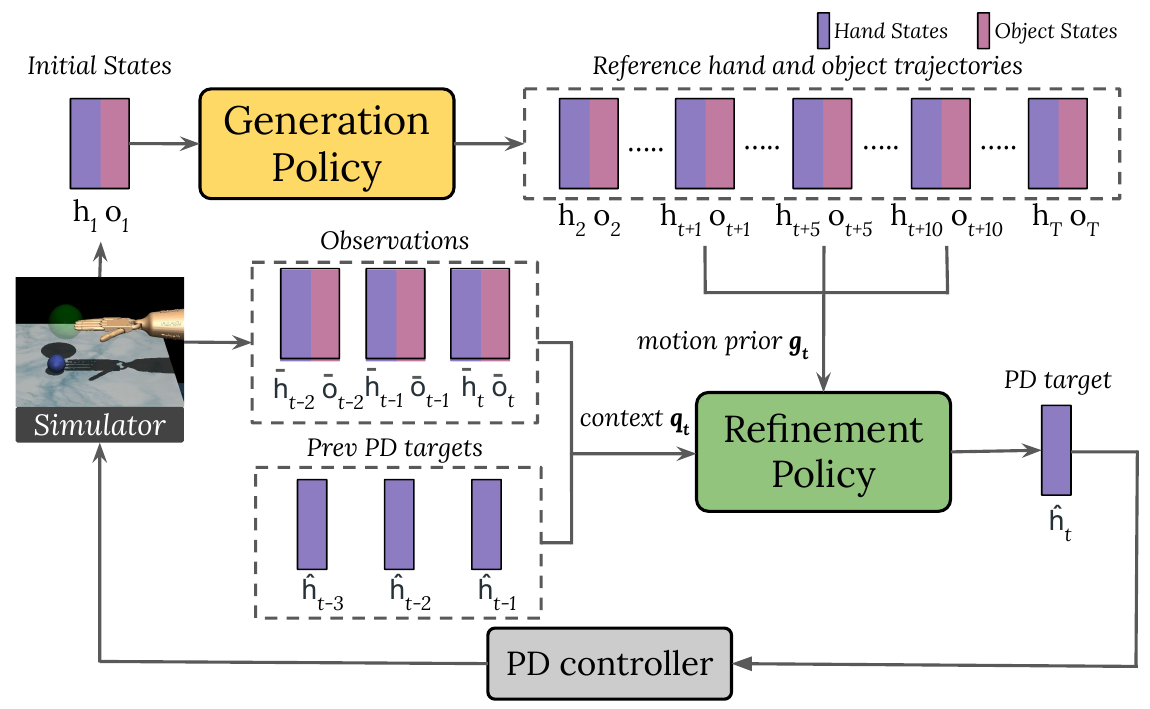}
     \caption{\edit{CIMER generates a motion prior based on initial conditions and refines it based on context to generate PD targets for the hand.}}
      \label{fig:data_flow}
\end{figure}

\textit{State \& Action Space}: We parameterize $\Psi$ as an MLP network that predicts the current action:  $\hat{\mathrm{h}}_t =\Psi(\mathrm{q}_t,\mathrm{g}_t)$. 
Here, $\hat{\mathrm{h}}(t) \in \mathcal{H}$ contains the refined PD targets of the robot hand at time $t$, 
$\mathrm{q}_t = (\bar{\mathrm{h}}_{t-2:t}, \bar{\mathrm{o}}_{t-2:t}, \hat{\mathrm{h}}_{t-3:t-1})$ is the \textit{context} containing the history of hand and object states and refined PD targets, and
$\mathrm{g}_t = (\mathrm{h}_{t+1,t+5,t+10}, \mathrm{o}_{t+1,t+5,t+10})$ provides the \textit{motion prior} (predicted by the motion generation policy $\Phi$) that needs to be refined. 
\edit{Since the necessary refinements are likely to be local and fine-grained, we restrict $\Psi$ to only refine the finger and palm trajectories.}
See Fig.~\ref{fig:data_flow} for an illustration.

\textit{Rewards}: Unlike prior works that rely on task-specific reward shaping, we use tracking errors of both the robot hand and the object as reward signals. 
Formally, we define the reward function as $r_t = r_t^{h} + r_t^{o} + r_t^{b}$, where
${r_t^{h} = \exp \left(-k^h || \bar{\mathrm{h}}_{t+1} - \mathrm{h}_{t+1}||^2 \right)}$, ${r_t^{o} = \exp \left(-k^o  || \bar{\mathrm{o}}_{t+1} - \mathrm{o}_{t+1}||^2 \right)}$,
$r_t^b$ is a bonus reward that is triggered when the object tracking error is smaller than a threshold $\epsilon_o$. Here, $\bar{\mathrm{h}}_{t+1}, \mathrm{h}_{t+1}$, and $\bar{\mathrm{o}}_{t+1}, \mathrm{o}_{t+1}$ denote the hand and object states from both the robot and reference motion, respectively.
Note that $r_t^h$ minimizes deviation from the reference hand motion $\mathrm{h}_{t+1}$ while $r_t^o$ encourages the robot hand to emulate the reference object motion $\mathrm{o}_{t+1}$. We include the training parameters in Appendix~\ref{appendix:CIMER_detail}.

\textit{Warm Start}: Since the motion generation policy $\Phi$ provides a reasonable motion prior, 
we expect the motion refinement policy $\Psi$ to only make local adjustments and account for contact effects. As such, we use the reference motions from $\Phi$ to warm start $\Psi$ by minimizing a behavior cloning loss. 
To counter covariate shift, we inject random noise into the policy's input and use the same outputs as corrective labels~\cite{nair2018overcoming, ke2021grasping, florence2019self}. This warm start represents yet another way in which imitation influences the emulation process (in addition to via the reward function and the policy input).


\begin{figure}
     \centering
    \includegraphics[width=1\linewidth]{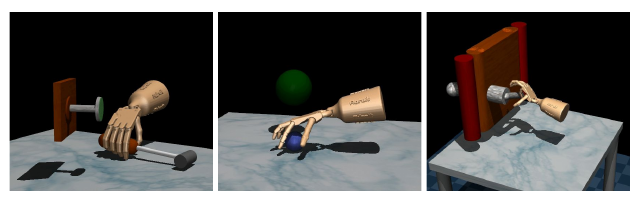}
     \caption{We evaluate on three dexterous prehensile skills from \cite{Rajeswaran-RSS-18}: Tool Use (left), Object Relocation (center), and Door Opening (right).}
      \label{fig:tasks}
\end{figure}
\section{Experimental Evaluation}
\edit{Our experiments pose the following questions: i) Is emulation necessary? (Sec.~\ref{subsec:why_emulation}), ii) Is CIMER more sample efficient than baselines? (Sec.~\ref{subsec:comparisons_baselines}), iii) Does CIMER generate robust and realistic motions? (Sec.~\ref{subsec:comparisons_baselines}), and iv) Can CIMER generalize and adapt to novel objects (Sec.~\ref{subsec:generalization}).}

\subsection{Experimental Design}
\label{subsec:exp_design}

\noindent\textbf{Evaluation Platform}: 
We used the widely-used ADROIT Hand~\cite{Rajeswaran-RSS-18} -- a 30-DoF simulated system (24-DoF hand + 6-DoF floating base) built with MuJoCo Simulator~\cite{Todorov2012MUJuCO}.

\noindent\textbf{Data and Expert Policy}:
We utilized expert DAPG policies~\cite{Rajeswaran-RSS-18} trained using action labels to generate 200 state-only observations (both hand and object states) for each task. 

\noindent\textbf{Tasks}: 
We consider three widely-used prehensile tasks originally proposed in \cite{Rajeswaran-RSS-18} (see Fig. \ref{fig:tasks} and Appendix~\ref{Appendix:state_design}).
\begin{itemize}
    \item \textit{Tool Use}: Pick up the hammer to drive the nail into the board placed at a randomized height.
    \item \textit{Object Relocation}: Move an object to a randomized target location (green sphere).
    \item \textit{Door Opening}: Given a randomized door position, unlock the latch and pull the door open.
\end{itemize}
\edit{To ensure fairness and reproducibility, we adopt the same success criteria originally proposed in \cite{Rajeswaran-RSS-18}.}

\noindent\textbf{Metrics}: 
We quantify performance in terms of \textit{Task success rate} (see~\cite{Rajeswaran-RSS-18, radosavovic2021state, han2023utility} for criteria) and sample efficiency, reported across five random seeds unless specified otherwise.

\begin{figure}[t]
     \centering
\includegraphics[width=0.99\linewidth]{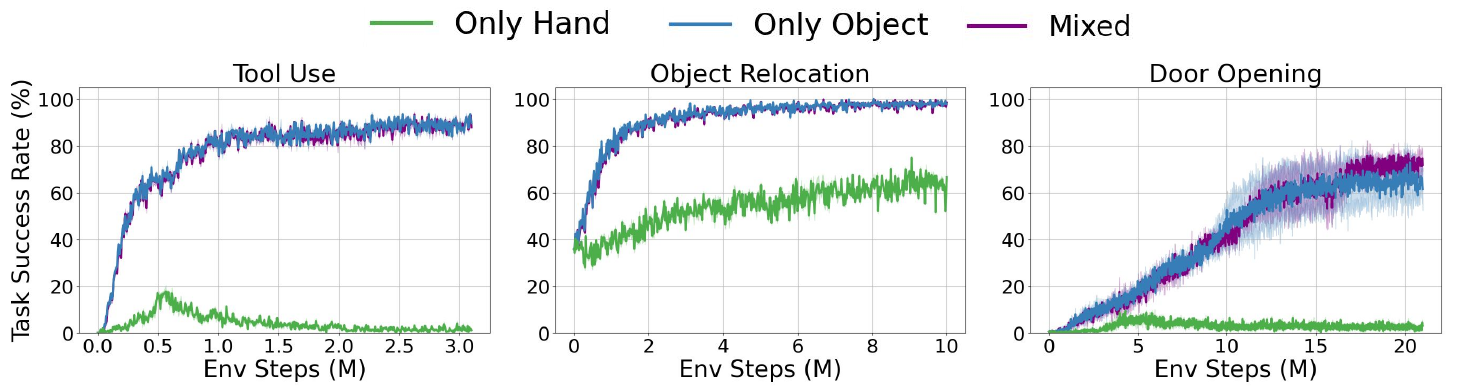}
     \caption{
     \edit{Emulating the observed object motion is significantly more effective than emulating the observed hand motion.}
     }
     
     \label{fig:rewards}
\end{figure}
\begin{figure}[t]
     \centering
\includegraphics[width=0.95\linewidth]{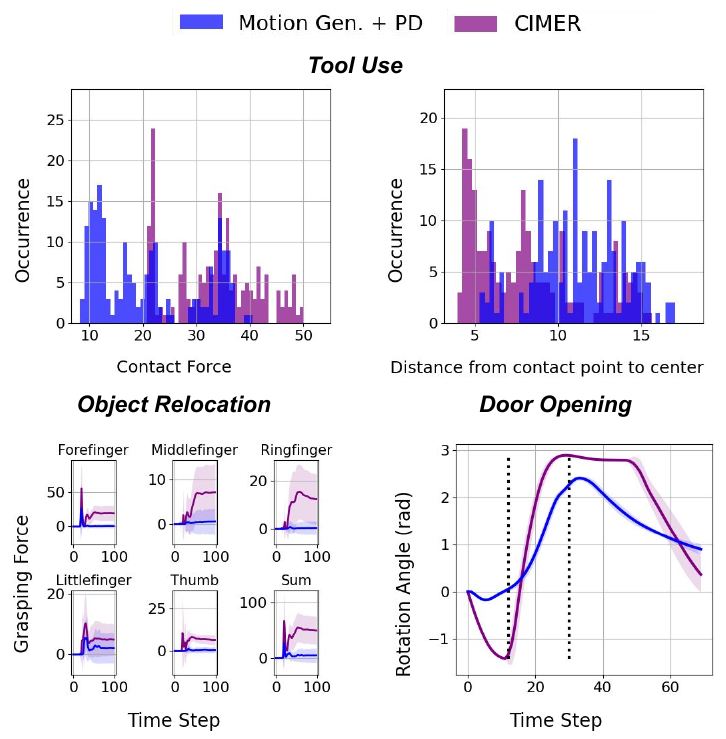}
     \caption{Intuitive refinements emerge from CIMER's emulation. \textit{Top}: Hand ensures hammer hits closer to nail's center, and applies larger driving force; \textit{Bottom Left}: Fingers exert more force to ensure firmer grasps and stable transport; \textit{Bottom Right}: Hand rotates faster to boost momentum while turning door handle (enclosed by dotted lines).}
     \label{fig:motion_ana}
\end{figure}
\begin{figure*}[t]
     \centering
\includegraphics[width=0.95\linewidth]{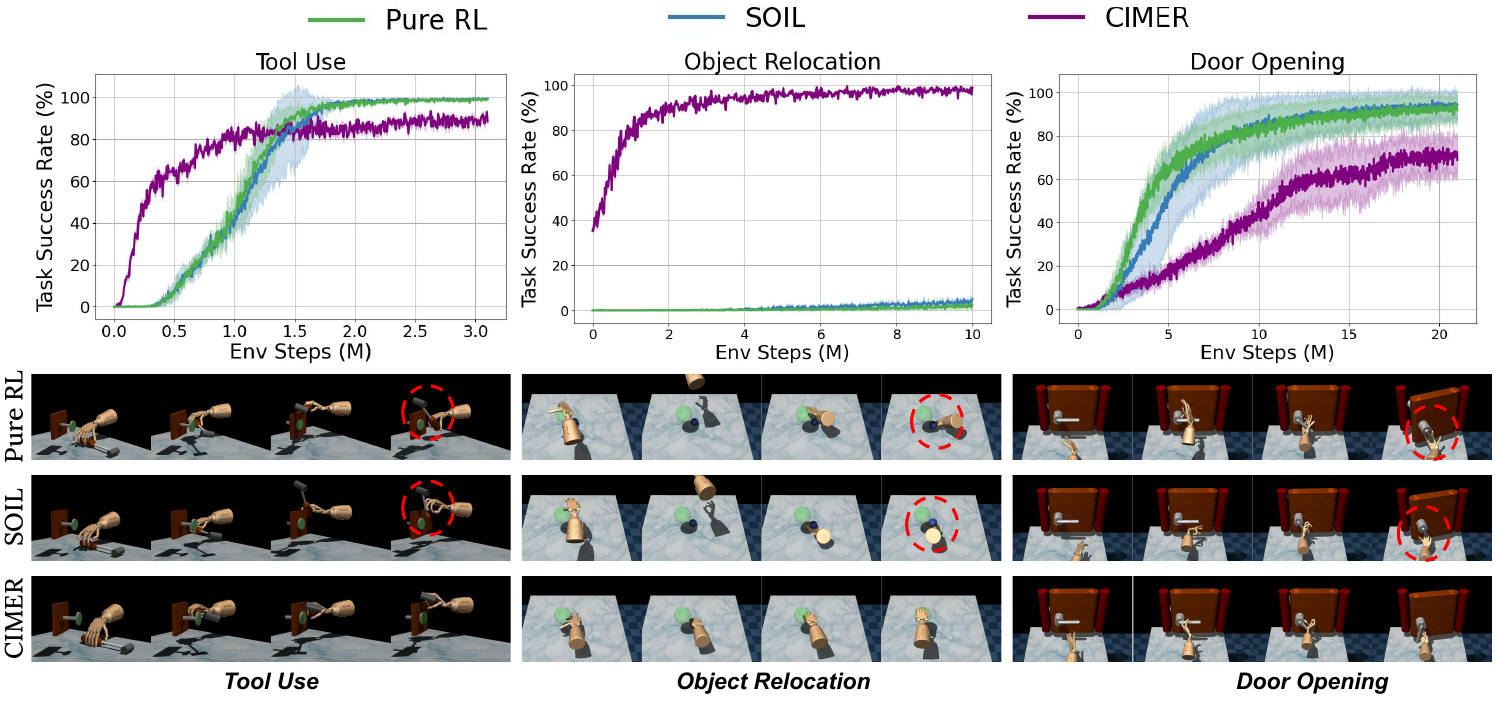}
    \caption{
    \textit{Top}: CIMER is considerably more sample efficient than the baselines in two tasks (\textit{Tool Use} and \textit{Object Relocation} tasks). The \textit{Door Opening} task is an exception likely due to larger refinements being necessary to firmly grasp and swing the door open.
    \textit{Bottom}: Qualitative analyses reveals that both the baselines exploit the simulator and generate unrealistic and aggressive behaviors in all three tasks (see dashed red circles indicating loss of grasp, bouncing objects, and hand-object penetration). In contrast, CIMER generates realistic motions as its refinement is anchored by the motion priors from its imitation phase.
    }
     \label{fig:soil_compare}
\end{figure*}

\subsection{\edit{Need for Emulation}}
\label{subsec:why_emulation}

\edit{We first demonstrate why it insufficient to imitate state-only observations without emulation. 
To this end, we compare CIMER against \edit{three} baselines that solely rely on imitation: a). \textit{Expert Obs.~\cite{Rajeswaran-RSS-18} + PD}: Reference hand motion generated by the expert policy trained with action labels, b). \textit{Motion Gen. + PD}: Reference hand motion generated by $\Phi$, and c). \textit{DexTransfer~\cite{chen2022dextransfer} + PD}: Reference hand motion generated by a policy trained on an augmented dataset (generated by simulating perturbed observations and augmenting new trajectories deemed successful by task-specific success criteria)}. We used the same PD controller with identical gains for all tasks and methods, and report success rates over 200 initial conditions.

\begin{table}[!htb]
    \caption{
    Tracking state trajectories from an expert or an imitation policy is insufficient for prehensile manipulation. Combing imitation and emulation (CIMER) dramatically improves task performance.
    }
    \centering
    \begin{tabular}{c|c|c|c}
    \hline
        \multirow{3}{*}{\diagbox[width=10.5em,height=3\line]{Policy}{Task Success}{Task}} & \multirow{3}{*}{Tool} & \multirow{3}{*}{Relocation} & \multirow{3}{*}{Door} \\ 
            Rate  &  & &  \\
            &  & &  \\
        \hline
    Expert~\cite{Rajeswaran-RSS-18} Obs. + PD   &  28.0\% & 34.5\% & 10.5\% \\
        \hline
    Motion Gen. + PD & 32.0\%  & 35.5\% & 10.5\%  
   \\ \hline
\edit{DexTransfer~\cite{chen2022dextransfer} + PD} & \edit{46.6($\pm$3.2)\%} & \edit{24.2($\pm$0.7)\%} & \edit{\edit{9.3($\pm$0.8)\%}} 
   \\ \hline
    CIMER + PD  & \textbf{92.7($\pm$1.7)}\%  & \textbf{96.0($\pm$0.5)}\% & \textbf{76.3($\pm$13.4)}\% \\ 
    \hline
    \end{tabular}
    \label{tab:PD_tracking}
\end{table}

\edit{As seen in Table \ref{tab:PD_tracking}, CIMER significantly outperforms all three baselines. These results support our claim that merely tracking an observed expert trajectory is insufficient for prehensile dexterous manipulation since contact effects and grasp forces are ignored when merely imitating state-only observations. 
Though DexTransfer uses the simulator to evaluate the effectiveness of perturbed trajectories, it only learns from successful trajectories and disregards ones that fail.
In contrast, CIMER uses emulation to ``practice" what it learned during imitation, refining the hand motion to reproduce the observed \textit{object} motion. 
We also observed that refining only the robot finger and palm motions may not always be sufficient. Specifically, we found that doing so improves task success to only 46.5($\pm$3.4)\% in the \textit{Door Opening} task. However, the success rate jumps to 76.3($\pm$13.4)\% when CIMER also refines the motion of the 6-DoF hand base. This more comprehensive refinement is likely helping the hand generate larger momentum to successfully turn the door handle.}

A key aspect of our approach to emulation is the increased focus on the object. To validate this focus, we evaluated the relative importance of hand- and object-tracking rewards. Results in Fig.~\ref{fig:rewards} highlight the crucial role of object tracking rewards. Importantly, using hand tracking rewards alone results in little to no performance gains.

To better understand how emulation improves performance, we qualitatively analyzed the modifications made by the motion refinement policy to compensate for the lack of action labels in each task (see Fig.~\ref{fig:motion_ana}). 
\textit{Tool Use}: The hammer hits the nail closer to its center point and with greater force (measured by a touch sensor on the nail head), helping drive the nail completely into the board. 
\textit{Object Relocation}: the grasping force applied from each fingertip on the object is noticeably increased, producing a more secure grasp during the initial contact and transportation towards the target. 
\textit{Door Opening}: the hand initially rotates in one direction, then swiftly changes direction to generate greater momentum for turning the handle (the dashed lines denote the handle turning phase). 
These observations suggest that emulation indeed nudges the robot to apply the necessary forces and adjusts the hand's motion so that the object is manipulated as desired. 

\subsection{\edit{Sample Efficiency and Realism}}
\label{subsec:comparisons_baselines}
\edit{Though no existing methods can accommodate all the constraints and challenges of our problem setting (see Sec.~\ref{subsec:formulation}), we compare CIMER to the following baselines to evaluate its relative benefits in terms of sample efficiency and realism:
\begin{itemize}
    \item \textit{Pure RL}: We trained TRPO~\cite{schulman2015trust} as a pure RL baseline without offline data, but with task-specific reward shaping~\cite{Rajeswaran-RSS-18} to evaluate the benefits of leveraging observations. 
    \item \textit{SOIL}~\cite{radosavovic2021state}: We also evaluated CIMER against this SOTA state-only imitation method for dexterous manipulation, which also leverages both expert observations and interactions with a simulator.
    Unlike CIMER, SOIL also requires task-specific reward engineering~\cite{Rajeswaran-RSS-18}.
\end{itemize}}
\noindent For both baselines, we utilized the same model architectures and training parameters as recommended in their original works. We report all results over five random seeds.

As shown in Fig.~\ref{fig:soil_compare} (top), CIMER exhibits significantly better sample efficiency than SOIL and Pure RL for both \textit{Tool Use} and \textit{Object Relocation} tasks. In particular, while the baselines struggle to make any progress on the \textit{Object Relocation} task, CIMER achieved 100\% task success rate. \edit{This is likely the baselines end up learning a fragile strategy of bouncing the ball against the table instead of grasping it (see Fig.~\ref{fig:soil_compare} (bottom))}. In contrast, baselines were able to achieve higher success rates in the \textit{Door Opening} task with fewer samples compared to CIMER. This is partly due to the door opening task likely requiring more significant motion refinements to generate enough momentum that will swing the door open. Further, unlike the baselines, CIMER does not benefit from task-specific reward design.

A closer examination revealed something important: the baselines tend to exploit the simulator and generate unrealistic or aggresive motions across all tasks. Exemplar rollouts in Fig.~\ref{fig:soil_compare} (bottom) show that the baselines tend to drive the hand to ``throw" the hammer, hit and bounce the ball, and penetrate the door handle, while CIMER generates more realistic and stable motions. This is because CIMER's motion generation policy employs a structured dynamical system that provides a realistic and effective motion prior for the motion refinement policy to adjust. These results also suggest that CIMER is better suited for real-world deployment and will not likely require significant efforts in ensuring safety and efficacy. 

\edit{
To quantitatively evaluate realism and the suitability for real-world implementation,
we evaluated the robustness of each method to systematic changes in all mass and damping coefficients (see Appendix~\ref{appendix:robustness} for details). The results reveal that CIMER results in comparable or better robustness, even outperforming the expert policy trained using action labels on the \textit{Relocation} task. Though Pure RL and SOIL seem to be more robust than CIMER on the \textit{Door} task, note that qualitative analysis revealed that both Pure RL and SOIL tend to exploit the simulator and generate unrealistic and agrressive motions (see Fig.~\ref{fig:soil_compare} (bottom)).}

\subsection{\edit{Generalization and Adaptation to Novel Objects}}
\label{subsec:generalization}

\begin{figure}[b]
     \centering
\includegraphics[width=0.99\linewidth]{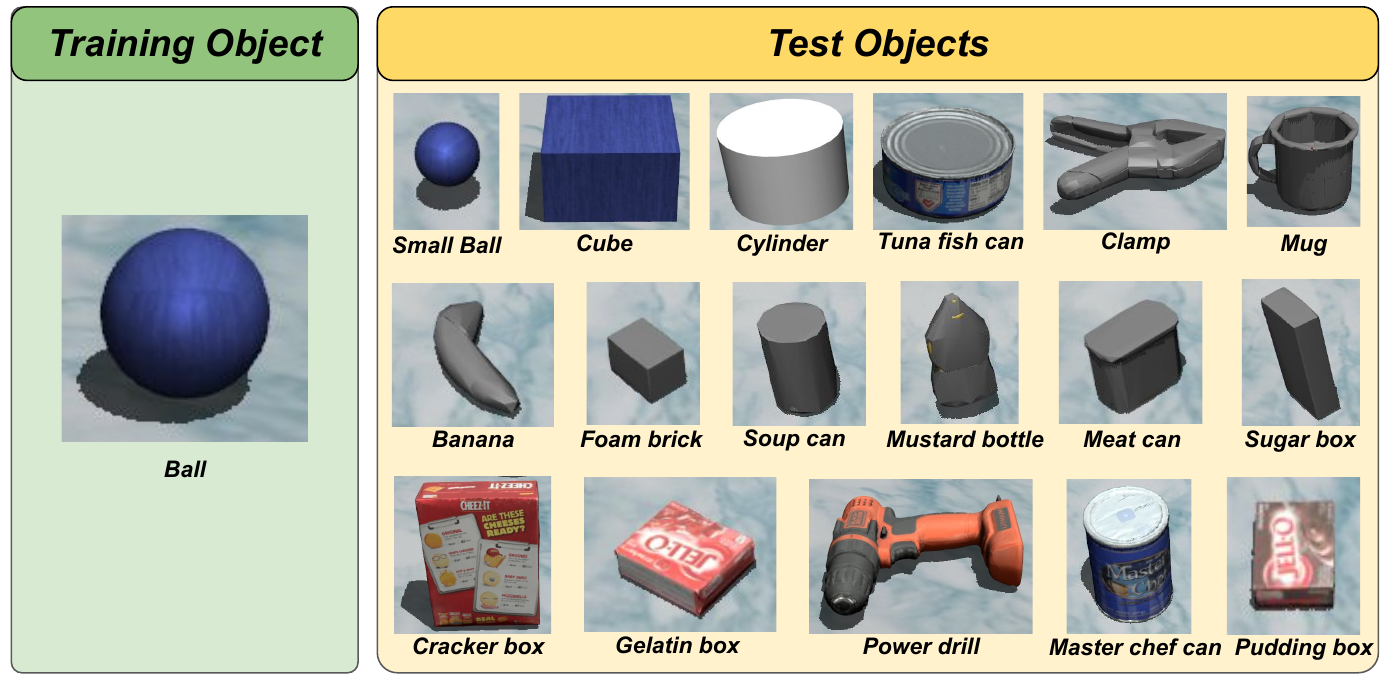}
     \caption{
     We trained candidate policies on one object, and evaluated their ability to generalize and adapt to 17 novel objects. 
     }
     \label{fig:objects}
\end{figure}

In addition to the objects considered thus far, we evaluated CIMER and the baselines on their ability to generalize to 17 novel objects (3 synthetic objects and 14 objects from the YCB dataset~\cite{calli2015benchmarking}) in the \textit{Object Relocation} task (see Fig.~\ref{fig:objects}). Note that the Ball is the only default object used for training the expert policy and for collecting state-only demonstrations.

\begin{figure}[t]
     \centering
\includegraphics[width=0.99\linewidth]{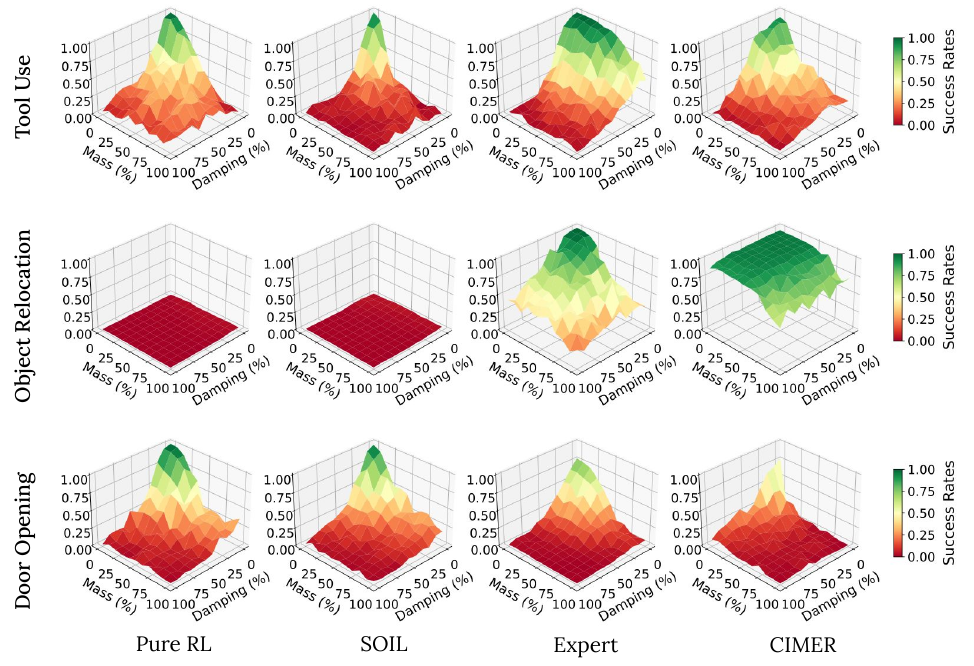}
    \caption{
     \edit{The evaluation of the robustness of each method to systematic changes in mass and damping coefficients.}
     }
     \label{fig:robustness_to_physics}
\end{figure}
\subsubsection{Zero-shot Generalization}

\begin{figure}[t]
     \centering
\includegraphics[width=0.95\linewidth,trim={0 11pt 0 5pt},clip]{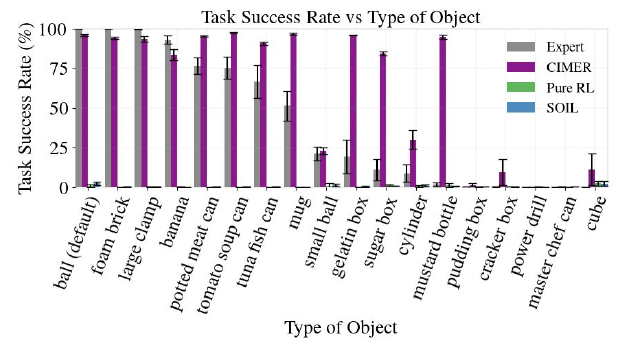}
     \caption{
     CIMER provides impressive zero-shot generalization to novel objects, even outperforming the expert policy in most cases.
     }
     \label{fig:zero}
\end{figure}

We report the zero-shot generalization performance of each method and the expert across five random seeds in Fig.~\ref{fig:zero}.
Given that both the baselines struggled to learn to relocate the default Ball object, it is no surprise that they consistently fail to generalize to novel objects. While the expert achieves a 100\% success rate on the default object (ball) with the benefit for action labels, CIMER notably outperforms the expert on the majority of the novel objects. This impressive ability to readily generalize to novel objects could be attributed to CIMER's hierarchical structure and suggests that CIMER's motion prior and refinement strategy are somewhat robust to changes in the object's geometric and kinematic properties. 
Further, CIMER demonstrates the most consistent performance across different training seeds, as indicated by shorter error bars. 

\subsubsection{Skill Transfer to New Objects}
We also evaluated if transferring the skills learned on the default object (Ball) to novel objects would improve sample efficiency
(see Appendix~\ref{appendix:object_adaptation} for experiments on learning from scratch). 
Freezing the motion generation policy, we fine-tuned CIMER's motion refinement policy on six novel objects (each of which resulted in 50\% or lower success rate during zero-shot generalization). We compared against a similarly-finetuned expert policy~\cite{Rajeswaran-RSS-18}, and two additional policies trained from scratch (one using CIMER and the other using the expert method).
\begin{figure}[t]
     \centering
\includegraphics[width=0.99\linewidth]{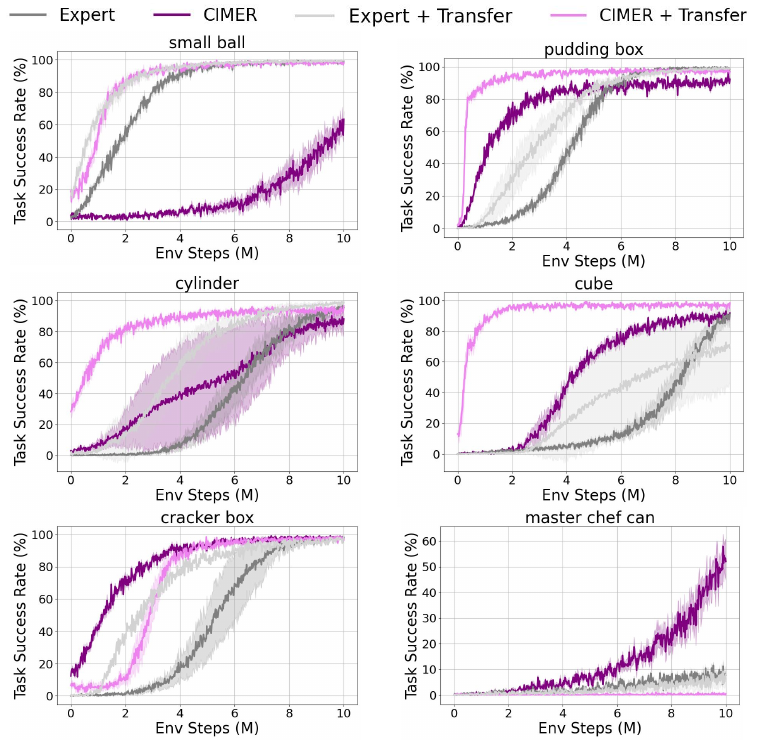}
     \caption{
     Skill transfer significantly boosts CIMER's sample efficiency (top two rows), unless the target object is considerably different from the source (bottom row).
     Solid lines show mean trends and shaded areas show $\pm$ standard deviation over five random seeds. 
     }
     \label{fig:sk_T}
\end{figure}
As shown in Fig.~\ref{fig:sk_T}, transferring motion refinement skills considerably improves sample efficiency in four of the six objects. On these four objects, CIMER with skill transfer outperforms the baselines trained from scratch as well as the fine-tuned expert. 
In contrast, skill transfer hurts sample efficiency when dealing with other two objects. This is likely because skill transfer is counter-productive when target objects are drastically different from the source object, requiring a significantly different grasping strategy. In such cases, CIMER has to ``unlearn" the previous skill before adapting to the novel object.

\section{Limitations and Future Work} 
\label{sec:limit}
Our work represents the first task-agnostic and intervention-free approach to learning dexterous prehensile manipulation skills from state-only observations. But it still leaves behind several avenues for further improvement. 
First, given that human videos are ubiquitous online, one could integrate recent advances in motion retargetting~\cite{qin2022dexmv} with CIMER to directly learning from videos. 
Second, we do not consider the manipulation of deformable and fragile objects. It is yet unclear how to represent such objects in way that makes incentivizing emulation easier and tractable.
Third, some objects and contexts require significantly different strategies (e.g., grasping a pencil to write vs. a ball to throw). It might be possible to enable such drastic adaptations by refining all available degrees of freedom of the hand and accounting for affordances. 
\edit{Fourth, reasoning about local contact events using tactile sensors~\cite{yuan2017gelsight, han2024learning, yu2023mimictouch} might help improve refinement and tackle contact-rich tasks (e.g., peg-in-hole).} 
Last, while our robustness experiments show promise, CIMER is yet to be implemented on physical robots. Leveraging the fact that it generates PD targets, we plan to deploy CIMER on hardware and validate its benefits.


\label{sec:limitation}
\bibliographystyle{ieeetr}
\bibliography{references}
\begin{appendices}
\section{State Space Designs}
\label{Appendix:state_design}
Our simulation ran at 500 Hz and the state design for each task is as follows:

\textit{Tool use:}
The floating hand base can only rotate along the $x$ and $y$ axes, resulting in $\mathrm{h}_t \in \mathcal{H} \subseteq \mathbb{R}^{26}$. Unlike other tasks, where the objects of interest are directly manipulated by the hand, this task requires indirect manipulation of the nail. As such, in addition to the hammer position, orientation, and corresponding velocities $\mathrm{p}^{\text{tool}}_t, \mathrm{o}^{\text{tool}}_t, \dot{\mathrm{p}}^{\text{tool}}_t, \dot{\mathrm{o}}^{\text{tool}}_t$ ($\mathbb{R}^{3}$), we define the nail goal position $\mathrm{p}^{\text{nail}}$
($\mathbb{R}^{3}$).
 Finally, we have $\mathrm{o}_t = [\mathrm{p}^{\text{tool}}_t, \mathrm{o}^{\text{tool}}_t, \dot{\mathrm{p}}^{\text{tool}}_t, \dot{\mathrm{o}}^{\text{tool}}_t, \mathrm{p}^{\text{nail}}] \in \mathcal{O} \subseteq \mathbb{R}^{15}$. We use the $\mathrm{p}^{\text{tool}}_t$ and $\mathrm{o}^{\text{tool}}_t$ as the object tracking states.

\textit{Object relocation:} The ADROIT hand is fully actuated, so we have $\mathrm{h}_t \in \mathcal{H} \subseteq \mathbb{R}^{30}$ (24-DoF hand + 6-DoF floating wrist base). Regarding the object states, we define $\mathrm{p}^{\text{target}}$ and $\mathrm{p}^{\text{ball}}_t$ as the target and current positions. Then, we compute $\bar{\mathrm{p}}^{\text{ball}}_t = \mathrm{p}^{\text{ball}}_t - \mathrm{p}^{\text{target}}$, which is the component of $\mathrm{p}^{\text{ball}}_t$ in a new coordinate frame that is constructed by $\mathrm{p}^{\text{target}}$ being the origin. 
 We additional include  the ball orientation $\mathrm{o}^{\text{ball}}_t$ and their corresponding velocities $\dot{\mathrm{p}}^{\text{ball}}_t$,  $\dot{\mathrm{o}}^{\text{ball}}_t$ (all $\mathbb{R}^{3}$).
Finally, we have $\mathrm{o}_t = [\bar{\mathrm{p}}^{\text{ball}}_t, \mathrm{o}^{\text{ball}}_t, \dot{\mathrm{p}}^{\text{ball}}_t, \dot{\mathrm{o}}^{\text{ball}}_t] \in \mathcal{O} \subseteq \mathbb{R}^{12}$. We use the $\bar{\mathrm{p}}^{\text{ball}}_t$ as the object tracking states.

\textit{Door opening:} For this task, the floating wrist base can only move along the direction that is perpendicular to the door plane but rotate freely, so we have $\mathrm{h}_t \in \mathcal{H} \subseteq \mathbb{R}^{28}$. Regarding the object states, we define the fixed door position $\mathrm{p}^{\text{door}}$, which can provide with case-specific information (similar to $\mathrm{p}^{\text{nail}}$ in \textit{Tool Use} task), and the handle positions $\mathrm{p}^{\text{handle}}_t$ (both $\mathbb{R}^{3}$). In order to take into consideration the status of door being opened, we include the angular velocity of the opening angle $v_t$$
 (\mathbb{R}^{1})$. Finally, we have $\mathrm{o}_t = [\mathrm{p}^{\text{handle}}_t, v_t, \mathrm{p}^{\text{door}}] \in \mathcal{O} \subseteq \mathbb{R}^{7}$. We use the $\mathrm{p}^{\text{handle}}_t$ as the object tracking states.





\section{CIMER Hyperparameters}
\label{appendix:CIMER_detail}
Table~\ref{tab:parameters} summarizes the hyper-parameter settings for training CIMER policies. For the three tasks, the hyper-parameter settings are the same.
 \begin{table}[!h]
    \caption{Training details}
    \centering
    \begin{tabular}{c|c|c}
    \hline
   MLP Hidden Layer    & RL Learning Rate & GAE$_\gamma$ \\
        \hline
    $(256, 128)$ & 2e-6  & 0.98 \\
        \hline
       PPO clip threshold    & PPO epochs  & GAE$_\lambda$ \\
        \hline
        0.2 & 8  & 0.97 \\
        \hline
    \end{tabular}
    \label{tab:parameters}
\end{table}

In addition, we used $k^h = 5, k^o = 5, r_t^b = 25$ in all tasks to compute the tracking rewards introduced in Sec.~\ref{subsec:emulation}.

\edit{
\section{Robustness Experiments}
\label{appendix:robustness}
For the robustness experiments in Sec.~\ref{subsec:comparisons_baselines}, we varied mass and damping since they are an important subset of differences between simulation and hardware. Specifically, we varied all mass and damping coefficients along eleven equally spaced evaluation points at $[0\%, 10\%, 20\%,  ..., 100\%]$ variation. For each variation, we sampled mass and damping coefficients from a uniform distribution, with the lower and upper bounds being multiples of the default values. The lower and upper bound values ($l_j$ and $u_j$) for each variation case ($j$) are 
\begin{equation*}
    \forall j \in \{0, 0.1, 0.2, ..., 1.0 \}, l_j = 1-0.8j, u_j =   1+4.0j.
\end{equation*}
After sampling a particular combination of mass and damping parameters, we tested each policy for 200 episodes with varying goal locations. 
We repeated this process 20 times for each variation, and report the resulting average success rates. 
}
\section{CIMER Pseudo-code}
\label{appendix:Pseudo_code}
The overall pseudo-code for CIMER is given below.
\begin{algorithm}
\SetAlgoLined
\nonl \textbf{Inputs:} State-only observation dataset $\mathcal{D}$, Koopman lifting function $g(\cdot)$; \\
\nonl \textbf{Motion Imitation} \\
Formulate Motion Generation Policy $\Phi$ as a Koopman-based dynamical system: $g({\mathrm{h}}_{t+1}, {\mathrm{o}}_{t+1}) = \mathbf{K}\ g({\mathrm{h}}_{t}, {\mathrm{o}}_{t})$ \\
Regress Koopman Matrix $\mathbf{K}$ on $\mathcal{D}$; \\
\nonl \textbf{Motion Emulation} \\
Warm-start the Motion Refinement Policy $\Psi$ to reconstruct trajectories generated by $\Phi$;\\
\For{$iter\in \{1, ..., \max\}$}{
Initialize replay buffer $\mathcal{B} = \varnothing$ ; \\
\For{$k\in \{1, ..., M\}$}{
Specify random initial states $\{{\mathrm{h}}_1^{(k)},{\mathrm{o}}_1^{(k)}\}$; \\
Generate reference motions $\{{\mathrm{h}}_t^{(k)},{\mathrm{o}}_t^{(k)}\}^{t=T^{(k)}}_{t=2}$ by rolling out $\Phi$; \\
\For{$t\in \{1, ..., T^{(k)}-1\}$}{
Motion Refinement Policy $\Psi$ inputs $\mathrm{s}_t$, and outputs $\mathrm{a}_t$ for execution; \\
Compute the tracking reward $r_t$;
}
Add $\{\mathrm{s}_t^{(k)}, \mathrm{a}_t^{(k)}, r_t^{(k)}\}^{t=T^{(k)}-1}_{t=1}$ to $\mathcal{B}$;\\
}
Update the Motion Refinement Policy $\Psi$ with $\mathcal{B}$; \\
}
\textbf{Return:} Trained policies $\Phi, \Psi$
\caption{CIMER}
\label{alg:pseudo-code}
\end{algorithm} 

\section{Details of KODex Training}
\label{appen:detail_kodex}
\textit{Modeling Dexterous Manipulation Skills}: 
A central principle behind KODex is that the desired behavior of a robot can be represented using a dynamical system. To this end, we define the state at time $t$ as $\mathrm{x}(t) = [{\mathrm{h}_t}^\top, {\mathrm{o}_t}^\top]^\top$, where ${\mathrm{h}_t} \in \mathcal{H} \subseteq \mathbb{R}^n$ and ${\mathrm{o}_t} \in \mathcal{O} \subseteq \mathbb{R}^m$ represent the state of the robot and the object, respectively, at time $t$. As such, the dynamical system we wish to capture is
\begin{equation}
\label{eqn:F*}
    \mathrm{x}(t+1) = F^*(\mathrm{x}(t))
\end{equation}
where $F^*(\cdot):\mathcal{H} \times \mathcal{O} \rightarrow \mathcal{H} \times \mathcal{O} $ denotes the dynamics that govern the \textit{interdependent} motions of the robot and the object.

A key challenge in learning the dynamical system in (\ref{eqn:F*}) is that it can be arbitrarily complex and highly nonlinear, depending on the particular skill of interest. KODex leverages Koopman operator theory to learn a \textit{linear} dynamical system that can effectively approximate such complex nonlinear dynamics. To this end, we first define the Koopman lifting function $g(\cdot)$ as follows
\begin{equation}
\label{eqn:gx}
g(\mathrm{x}(t)) = [{\mathrm{h}_t}^\top, \psi_h({\mathrm{h}_t}),  {\mathrm{o}_t}^\top, \psi_o({\mathrm{o}_t})]^\top, \forall t
\end{equation}
where $\psi_h:\mathbb{R}^n \rightarrow \mathbb{R}^{\bar{n}}$ and $\psi_o:\mathbb{R}^m \rightarrow \mathbb{R}^{\bar{m}}$ are vector-valued lifting functions that transform the robot and object state respectively. 

Now, we introduce the Koopman matrix $\mathbf{K}$, which approximates the system evolution in the lifted state space as follows
\begin{equation}
\label{eqn:approximate_KG}
 g(\mathrm{x}(t+1)) =  \mathbf{K}g(\mathrm{x}(t))
\end{equation}

\textit{Learning the Koopman Matrix $\mathbf{K}$}:
The next step is to learn the Koopman matrix $\mathbf{K}$ from dataset $\mathcal{D} = [\{{\mathrm{h}}_t^{(1)}, {\mathrm{o}}_t^{(1)}\}^{t=T^{(1)}}_{t=1}, \cdots, \{{\mathrm{h}}_t^{(N)},{\mathrm{o}}_t^{(N)}\}^{t=T^{(N)}}_{t=1}]$ (Section.~\ref{subsec:overview}). Recall that CIMER's Motion Generation Policy $\Phi$ is formulated as the linear dynamical system shown in~(\ref{eqn:approximate_KG}). 

As described in \cite{han2023utility}, we can efficiently compute the Koopman matrix as $\mathbf{K} = \mathbf{A}\mathbf{G}^\dagger$, where $\mathbf{A}$ and $\mathbf{G}$ are shown as follows
\begin{equation}
\label{eqn:k_comp_new}
\begin{split}
\mathbf{A} &=\sum_{n=1}^{n=N}\sum_{t=1}^{t=T^{(n)}-1}  \frac{g(\mathrm{x}^{(n)}(t+1)) \otimes g(\mathrm{x}^{(n)}(t))}{N(T^{(n)}-1)}, \\
\mathbf{G} &= \sum_{n=1}^{n=N}\sum_{t=1}^{t=T^{(n)}-1} \frac{g(\mathrm{x}^{(n)}(t)) \otimes g(\mathrm{x}^{(n)}(t))}{N(T^{(n)}-1)}   
\end{split}
\end{equation}
where $\mathbf{G}^\dagger$ denotes the Moore–Penrose inverse\footnote{It could be efficiently computed using  the\texttt{scipy.linalg.pinv($\mathbf{G}$)} function from Scipy library.} of $\mathbf{G}$, and $\otimes$ denotes the outer product. 

We use the computed Koopman matrix $\mathbf{K}$ to generate rollouts in the lifted state space. However, we need to obtain the rollouts in the original state states to retrieve the predicted robot and object states. Since we designed $g(\mathrm{x}(t))$ such that robot state $\mathrm{h}_t$ and the object state $\mathrm{o}_t$ are parts of lifted states in  (\ref{eqn:gx}), we can easily retrieve the reference motions of both the robot and the object $\{\mathrm{h}_t, \mathrm{o}_t\}$ by selecting the corresponding elements in $g(\mathrm{x}(t))$. \

\section{Emulation might not be necessary for Non-prehensile Manipulation}
\label{appendix:non_prehensile}
We present the task success rate on another non-prehensile manipulation task, \textit{In-hand Reorientation}, which are as follows: \textit{Expert Obs. + PD}: 91.0\%, \textit{Motion Gen. + PD}: 69.5\%, and \textit{CIMER}: 72.7($\pm$1.0)\%. Contrary to the results presented in Table.~\ref{tab:PD_tracking}, where \textit{Expert Obs. + PD} shows poor performance on three dexterous prehensile manipulation tasks, in this scenario, it could achieves satisfactory results. This disparity arises because the challenge of the \textit{In-hand Reorientation} task primarily pertains to the generation of precise and diverse hand motions, rather than dealing with heightened sensitivity to applied forces. Therefore tracking the observed expert motion directly translates to achieving the desired object motion. This is also why both the \textit{Motion Gen. + PD} and \textit{CIMER} are not comparable to \textit{Expert Obs. + PD} in this \textit{In-hand Reorientation} task. Therefore, we exclusively apply CIMER policies to the three dexterous prehensile manipulation tasks, aligning with the intended purpose of motion refinement policies. We leave the improvement of motion generation policy for non-prehensile tasks as a topic for future work.

\section{Adaptation to New Objects}
\label{appendix:object_adaptation}
We also evaluated CIMER's ability to adapt to novel objects. 
Specifically, we finetuned the CIMER on each of the seven objects 
for which its success rates was lower than 50\%. 
To ensure that CIMER remains intervention-free, we did not collect or rely on additional observations. Instead, we reused the old motion generation policy (trained on the default object), and simply changed the object used in the simulator when training the motion refinement policy. 
We similarly trained the expert policies to ensure a fair comparison. 

\begin{figure*}[!htb]
     \centering
\includegraphics[width=0.99\linewidth]{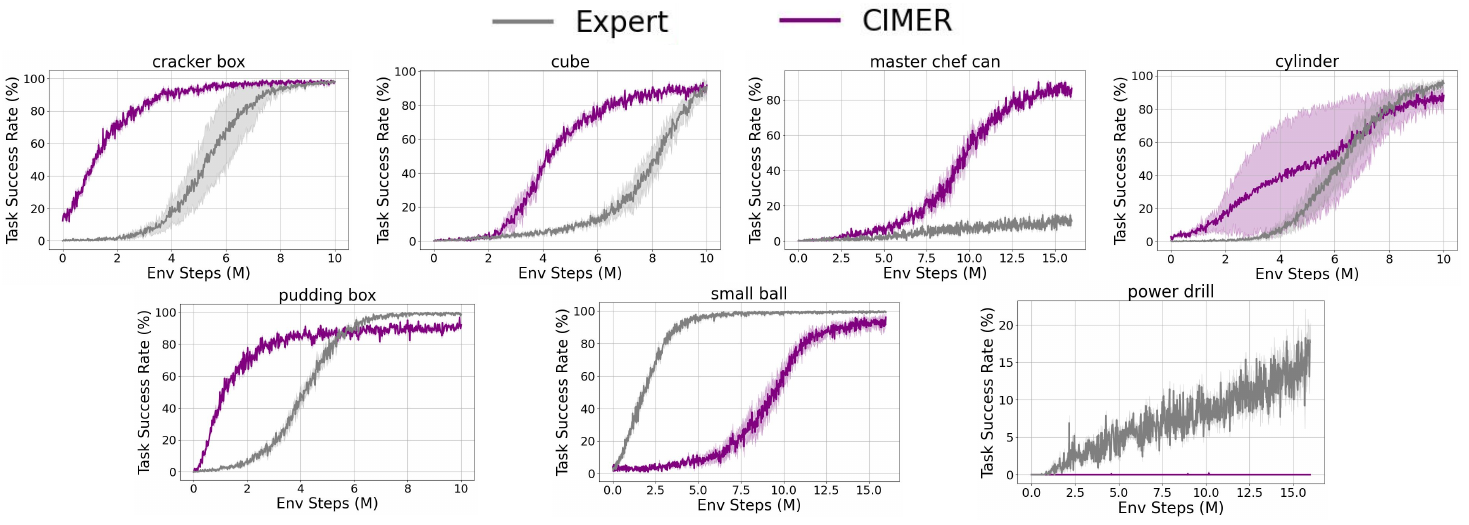}
     \caption{We compare the RL sample efficiency between CIMER and expert policies when being trained on new objects. For each, solid lines indicate mean trends and shaded areas show $\pm$ standard deviation over five random seeds. We  find that both methods will gradually learn the manipulation strategy of the new objects, except for the power drill. Moreover, CIMER policies exhibit the better sample efficiency.}
     \label{fig:fine-tuning}
\end{figure*}

As shown in Fig.~\ref{fig:fine-tuning}, we find that both methods gradually learn to manipulate the new objects, with CIMER exhibiting significantly better sample efficiency almost all objects. The power drill presents a notable exception to this trend, with CIMER particularly struggling to adapt.
This is due to the power drill likely requiring a significantly different grasping strategy than the default object (Ball) on which CIMER's motion generation policy was trained. Indeed, CIMER's motion refinement only adjusts finger and palm motions, and does not account for grasp synthesis. This could potentially be addressed by also learning to refine the hand base motions with the help of appropriate regularizers~\cite{cheng2019control} that encourage refined motions to remain close to the reference motion.

\end{appendices}

\end{document}